\documentclass[]{spie}  

\usepackage{amsmath,amsfonts,amssymb}
\usepackage{graphicx}
\usepackage[colorlinks=true, allcolors=blue]{hyperref}
\usepackage{picinpar}
\usepackage{longtable}

\usepackage{ifthen}
\usepackage[table]{xcolor}
\def\naf{ \: not \: } 
\definecolor{Gray}{rgb}{0.47,0.53,0.6}
\definecolor{LightCyan}{rgb}{0.88,1,1}
\definecolor{yellow}{rgb}{1,1,0.88}
\definecolor{azure}{rgb}{0.94,1,1}
\definecolor{lightgrey}{rgb}{0.83,0.83,0.83}
\newcommand{\uffa}{\mbox{$\: {\tt : \!\! - }\:$}} 

\usepackage{xcolor}
\usepackage[linesnumbered,ruled,vlined]{algorithm2e}

\SetCommentSty{mycommfont}

\SetKwInput{KwInput}{Input}                
\SetKwInput{KwOutput}{Output}              
\SetKw{Break}{break}
\SetKw{Go}{goto line}
\usepackage{multicol}

\usepackage{listings}
\lstdefinelanguage{clingo}{
  keywordstyle=[1]\usefont{OT1}{cmtt}{m}{n},%
  keywordstyle=[2]\textbf,%
  keywordstyle=[3]\usefont{OT1}{cmtt}{m}{n},
  alsoletter={\#,\&},%
  keywords=[1]{not,from,import,exists,if,else,return,while,
		break,and,or,for,in,del,and,class,subClass,concern,aspect,subCo,prop,rdf,cpsf,addBy,suppBy,neg,d,relation,holds,h,obs,action,fluent,occurs,req,group,leadTo,active,deg,comp,order,hSubCo,llh_sat_sub,llh_sat,llh_sat_sub_aux,step,deg_pos,nAllPosCon,nActPosCon,possImpactsPos,scoreLoS,wBus,wHum,wTru,wFun,wTim,wBou,wLif,wCom,wDat,last,addFun,func,formula,sat_formula,possImpactsNeg,conflict,member,sa_action,exec,sc_concern},%
  keywords=[2]{\#const,\#show,\#minimize,\#maximize,\#base,\#theory,\#count,\#external,\#program,\#script,\#end,\#heuristic,\#edge,\#project,\#show},%
  keywords=[3]{&,&dom,&sum,&diff,&show,&minimize},%
  morecomment=[l]{\#\ },%
  morecomment=[l]{\%\ },%
  commentstyle={\color{darkgray}}%
}
\def\easp{{\tt exp(ASP)}}

\newtheorem{example}{Example}  
 
\title{Generating explanations for answer set programming applications}

\author[a]{Ly Ly Trieu}
\author[a]{Tran Cao Son}
\author[a]{Enrico Pontelli}
\author[b]{Marcello Balduccini}  
\affil[a]{New Mexico State University, MSC CS, PO Box 30001, Las Cruces, New Mexico, USA}
\affil[b]{Saint Joseph's University, 5600 City Avenue, Philadelphia, PA, USA}

\begin{document} 
\maketitle

\begin{abstract}
 We present an explanation system for applications that leverage Answer Set Programming (ASP). 
 Given a program $P$, an answer set $A$ of $P$, and an atom $a$ in the program $P$, our system generates  all explanation graphs of $a$ which help explain why $a$ is true (or false) given the program $P$ and the answer set $A$.  
 We illustrate the functionality of the system using some examples from the literature.
\end{abstract}


\section{INTRODUCTION}
\label{sec:intro} 
\begin{figwindow}[2,r,%
{
\includegraphics[width=0.45\textwidth]{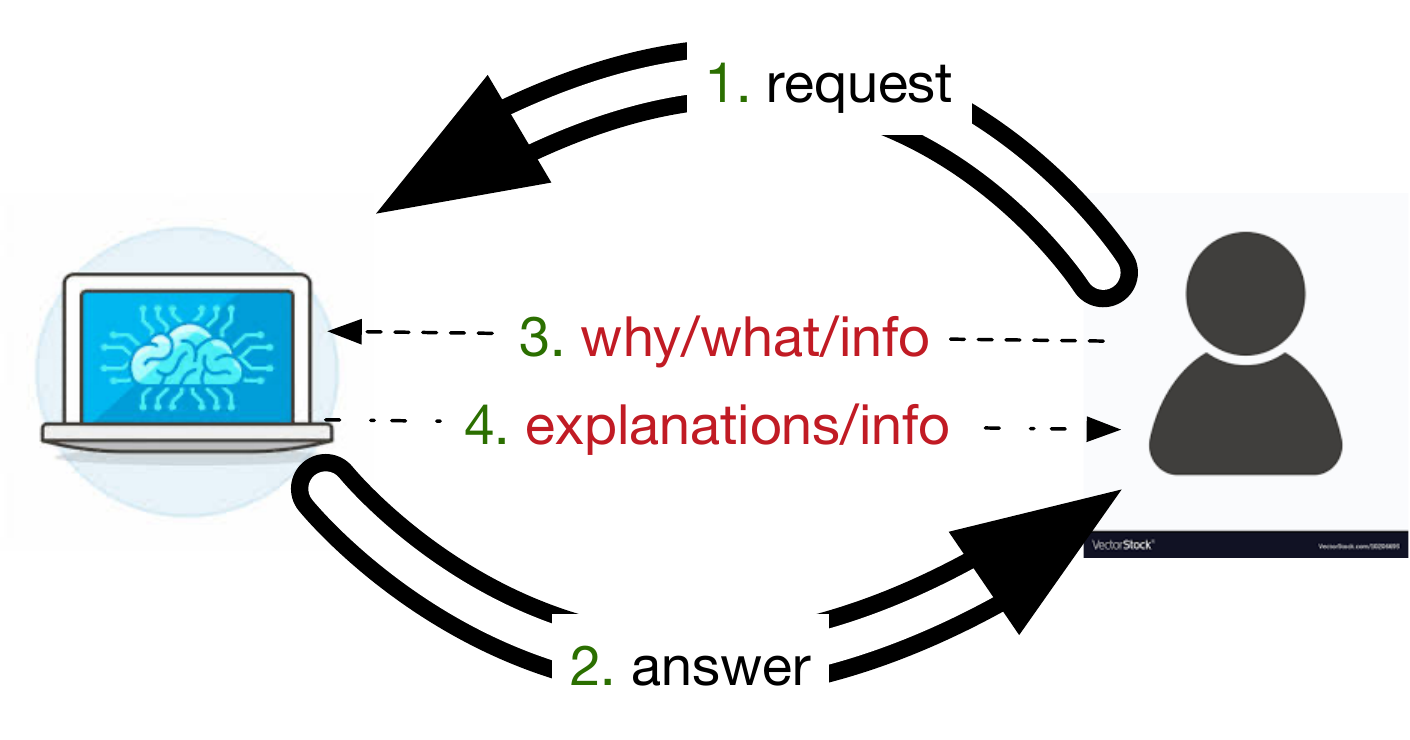} 
},%
{\scriptsize Interactions between AI System and Human User \label{fig0}}] 
\noindent In recent years, explainable AI has been introduced to help  users gain confidence in the AI's decisions and conclusions. We aim at developing AI systems capable of explaining their responses to a request from a user until the user finds that the answers are acceptable as illustrated in Fig.~\ref{fig0}. It starts with the user sending a request (Step 1). The system responds with an initial answer (2). Thereafter, a dialog between the two ensues with user sending questions and information---that the user believes  might be different than those owned by the system---to the system (3), and the system responds with explanations and information---that the system believes  the user might not have or have in incorrect form (4). This process continues until the  user agrees with the AI system. During this process, the AI will need to take into consideration the human model of the situation and the information sent by the user, differentiate it from its own, and inform the user about missing or false information. In addition, the AI's model of the human will need to be learned and updated as exchanges between the two sides occur. Most current AI systems lack components for dealing with the steps (3)-(4). 
\end{figwindow} 

\hspace{4mm} In this paper, we focus on a key component needed in steps (3)-(4) in the above architecture. We propose  an \emph{explainable Answer Set Programming (ASP)} system. ASP is a programming paradigm that has been applied in several applications such as planning, diagnosis, robotics, etc. ASP is attractive as it is declarative, non-monotonic, and elaboration tolerant, and has free and scalable solvers. Thus far, only limited attention has been paid to explaining the output of an ASP execution. The proposed system will take a program $P$ and a model $A$ of $P$ and explain ``why an atom is true/false in $A$.'' In addition, the system will also produce explanation graphs for atoms in an ASP program.  

The paper is organized as follows. We will start with a short illustration of answer set programming with focus on why ASP is an appropriate paradigm for the development of AI applications and when is it most appropriate for using ASP. Afterwards, we will define the notion of a justification (explanation graph) of an atom in a given answer set. We will then describe a system for generating explanations for answer set programming and present some initial applications. 

\section{PRELIMINARIES}

\subsection{Answer Set Programming (ASP)}
\label{sec:asp}

(ASP)~\cite{MarekT99,Niemela99} is a declarative programming paradigm based on 
logic programming under the answer set semantics.   
A logic program $P$ is a set of rules of the form  \quad 
\begin{equation}\label{rule}
c  \leftarrow a_1,\ldots,a_m,\naf b_{1},\ldots,\naf b_n 
\end{equation}
where $c$, $a_i$'s, and $b_j$'s are atoms of a propositional language\footnote{For simplicity, we often use first order logic atoms, as a representation of all of their ground instantiations.} and 
$\mathit{not}$ represents (default) negation. 
Intuitively, a rule  states that if all $a_i$ are  believed to be true and none of the $b_j$ is believed to be true then $c$ will be true. For a rule $r$,  
$r^+$ and $r^-$   denote the sets $\{a_1,\ldots,a_m\}$ and $\{b_{1},\ldots,b_n\}$, respectively.  We write $head(r)$ and $body(r)$ to denote $c$ and the right side of a  rule $r$. Both the head and the body or a rule $r$ can be empty; when the body is empty, the rule is called a fact; when the head is empty, it is a constraint. 
We use $H$ to denote the Herbrand base of a logic program $P$, which is the set of all ground atoms in $P$. 
 
Let $P$ be a program. An interpretation $I$ of $P$ is a 
subset of $H$. $I$ satisfies an atom $a$ ($I\models a$) if $a\in I$.
The body of a rule $r$ is satisfied by $I$ if $r^+ \subseteq I$ and $r^- \cap I = \emptyset$.
A rule $r$ is satisfied by $I$ if $I\models head(r)$ or $I \not\models body(r)$.  
$I$ is a model of $P$ if it satisfies all rules in $P$. 

For an interpretation
$I$ and a program $P$, the \emph{reduct}
of $P$ w.r.t. $I$ (denoted by $P^I$) is the program
obtained from $P$ by deleting
{\em (i)} each rule $r$ such that $r^- \cap I \neq \emptyset$, and
{\em (ii)} all elements of the form $\naf a$ in the bodies of the remaining rules.  
Given an interpretation $I$,
observe that the program $P^I$ is a program with no occurrences of $\naf a$. An interpretation $I$ is an 
\emph{answer set} of $P$ if $I$ is the least model (w.r.t. $\subseteq$) of $P^I$
\cite{GelfondL90}.  
Answer sets of logic programs can be computed using efficient and scalable answer set solvers, 
such as \textit{clingo}\cite{GebserKNS07}.

We illustrate the concepts of answer set programming by
showing how the 3-coloring problem of a undirected graph $G$ can
be solved using ASP.  Let the three colors be red ($r$),
blue ($b$), and green ($g$), and the vertices of $G$ be
$0, 1, \ldots, n$. Let $P(G)$ be the program consisting of

\begin{itemize}
\item the set of atoms $edge(u,v)$ and $edge(v,u)$ for every pair of connected vertices $u,v$ in $G$,

\item for each vertex  $u$ of $G$, three rules stating that $u$ must
be assigned one of the colors red, blue, or green ($colored(x,c)$ denotes 
that the node $x$ is colored with the color $c$):
\begin{eqnarray*}
colored(u, g) \leftarrow \naf colored(u, b), \naf colored(u,r) \\
colored(u, r) \leftarrow \naf colored(u, b), \naf colored(u,g) \\
colored(u, b) \leftarrow \naf colored(u, r), \naf colored(u,g)
\end{eqnarray*}
and

\item for each edge $(u,v)$ of $G$, three rules representing the
constraint that $u$ and $v$ must have different color:
\begin{eqnarray*}
 \leftarrow colored(u, r), colored(v, r), edge(u,v) \\
 \leftarrow colored(u, b), colored(v, b), edge(u,v) \\
 \leftarrow colored(u, g), colored(v, g), edge(u,v)
\end{eqnarray*}
\end{itemize}
It can be shown that for each graph $G$, (i) $P(G)$ does not have answer sets iff the 3-coloring problem of $G$ does not have a
solution; and (ii) if $P(G)$ has solutions,  then each answer set of
$P(G)$ corresponds to a solution of the 3-coloring problem of $G$
and vice versa.

\subsection{Properties of ASP}
\label{sec:properties_asp}

Given a program $P$ and an answer set $A$, the following properties hold:
\begin{enumerate}
    \item If $c \in A$, there exists a rule $r$ in $P$ such that  
        \begin{itemize}
            \item $head(r) = c$; 
            \item $r^+ \subseteq A$; and,
            \item $r^- \cap A = \emptyset$
        \end{itemize}
        For such a rule $r$, we define $support(c,r) = \{p \mid p \in A \land p \in r^+\} \cup \{\sim\!n \mid n \notin A \land n \in r^-\}$ and refer to this set as a \emph{supported set} of $c$ for rule $r$. 
    \item If $c \notin A$, for every rules $r$ such that $head(r)=c$, then  

        \begin{itemize}
            \item $r^+ \setminus A \ne \emptyset$; or,
            \item $r^- \cap A \ne \emptyset$
        \end{itemize}
        For such a rule $r$, we define $support(\sim\!c,r) \in \{\{p\} \mid p \in A \land p \in r^- \} \cup \{\{\sim\!n \} \mid n \notin A \land n \in r^+\}$ and refer to it as a supported set of $\sim\!c$ for rule $r$. 
\end{enumerate}

\subsection{Explanation Graph}
\label{sec:explanation_graph}

We rely on the notion of an off-line justification introduced by \cite{PontelliSE09} as an explanation of an atom $a$ given an answer set $A$ of a program $P$. $NANT(P) = \{ a \mid a \in  r^- \land  r \in P\}$ is the set of all negation atoms in $P$. Let $C(P)$ denote the set of cautious consequences of $P$, i.e., $C(P) = C^+ \cup C^-$  where  $C^+$ is  the set of atoms belonging to all answer sets of $P$ and $C^-$ is the set of atoms which do not belong to any answer set of $P$. A set of atoms $U$ such that $U \subseteq NANT(P) \setminus (A \cup C(P))$ is called a set of assumptions with respect to $A$ if $A = C(P \setminus \{r \in P \mid head(r) \in U\})$. 
  
Intuitively, given an answer set $A$ of a program $P$, an atom $a \in A$ ($a \not\in A$) is considered to be true (false) given $A$. An off-line justification for an atom $a$ presents a possible reason for the truth value of $a$, i.e., it answers the question ``\emph{why $a \in A$  (or $\not\in A$)?}''. If $a$ is true in $A$, an off-line justification of $a$ represents a derivation of $a$ from the set of assumptions $U$ and the set of facts in $P$. If $a$ is false in $A$, an off-line justification encodes the reason why it is not supported by $A$, which can be that it is assumed to be false (being an assumption in $U$) or there exists no possible derivation of it given $A$. An off-line justification of $a$ is analogous to the well-known SLDNF tree in that it represents the derivation for $a$. The key difference between these two notions is that an off-line justification might contain a cycle consisting of negative atoms, i.e., atoms not belonging to $A$. For this reason, a justification is represented as an explanation graph, defined as follows.   

\begin{definition}
[Explanation Graph]
\label{def:egraph}
Let us consider a program $P$, an answer set $A$, and a set of assumptions $U$ with respect to $A$. 
Let $N = \{x \mid  x \in A\} \cup \{ \sim{x} \mid x \not\in A\} \cup \{\top , \bot, \mathtt{assume}\}$ where $\top$ and $\bot$ represent true and false, respectively. 
An \emph{explanation graph} of an atom $a$ occurring in $P$ is a finite  labeled and directed 
graph $DG_a=(N_a,E_a)$ with $N_a \subseteq  N$ and $E_a \subseteq N_a \times N_a \times  \{+,-,\circ\}$, where $(x,y,z) \in E_a$ 
represents a link from $x$ to $y$ with the label $z$,  and satisfies the following conditions:  
\begin{itemize} 
\item if $a \in A$ then $a \in N_a$ and every node in $N_a$ must be reachable from $a$;  
\item if $a \not\in A$ then $\sim\!a \in N_a$ and every node in $N_a$ must be reachable from $\sim\!a$; 
\item if $(x, \top,+) \in E_a$ then $x$ is a fact in $P$;  
\item if $(\sim\!x, \mathtt{assume}, \circ) \in E_a$ then $x \in U$; 
\item if $(\sim\!x, \bot, +) \in E_a$ then there exists no rule in $P$ whose head is $x$;
\item there exists no $x,y$ such that $(\top, x, y) \in E_a$, $(\bot, x, y) \in E_a$, or $(\mathtt{assume}, x, y) \in E_a$;
\item for every $x \in N_a \cap A$ and $x$ is not a fact in $P$, 
	\begin{itemize} 
	\item there exists no $y \in N_a \cap A$ such that  $(x,y,-)$ or $(x,y,\circ)$ belong to $E_a$; 
	\item there exists no $\sim\!y \in N_a \cap \{\sim\!u \mid u \not\in A\}$ such that  $(x,\sim\!y,+)$ or $(x,\sim\!y,\circ)$ belong to $E_a$; 
	\item if $X^+ = \{a \mid (x, a,+) \in E_a\}$ and $X^- = \{a \mid (x, \sim\!a,-) \in E_a\}$ then 
	$X^+ \subseteq A$,  $X^- \cap A = \emptyset$, and there is a rule $r\in P$ whose head is $x$ such that  
	$r^+ = X^+$ and $r^- = X^-$; and  
	\item $DG_a$ contains no cycle containing $x$.  
	\end{itemize} 
\item for every $\sim\!x \in N_a \cap \{\sim\!u \mid u \not\in A\}$ and $x \not\in U$, 
	\begin{itemize} 
	\item there exists no $y \in N_a \cap A$ such that  $(\sim\!x,y,+)$ or $(\sim\!x,y,\circ)$ belong to $E_a$; 
	\item there exists no $\sim\!y \in N_a \cap \{\sim\!u \mid u \not\in A\}$ such that  $(\sim\!x, \sim\!y,-)$ or $(\sim\!x, \sim\!y,\circ)$ belong to $E_a$; 
	\item if $X^+ = \{a \mid (\sim\!x,   a,-) \in E_a\}$ and $X^- = \{a \mid (\sim\!x,  \sim\!a,+) \in E_a\}$ then 
	$X^+ \subseteq A$,  $X^- \cap A = \emptyset$, and for every rule $r\in P$ whose head is $x$ we have that
	 $r^+ \cap X^- \ne \emptyset$ or 	$r^- \cap X^+ \ne \emptyset$; and  
	\item any cycle containing $\sim\!x$ in $DG_a$ contains only nodes in $N_a \cap \{\sim\!u \mid u \not\in A\}$.  
	\end{itemize} 
\end{itemize} 
\end{definition}

Given an explanation graph $G$ and a node $x$ in $G$, if $x$ is an atom $a$ then the nodes directly connected to $a$---the nodes $y$ such that $(x,y,\_)$ is an edge in $G$---represent a rule whose head is $x$ and whose body is satisfied by $A$. If $x$ is $\sim\!x$ for some atom $x$,
then the set of nodes directly connected to $\sim\!x$ represents a set of atoms who truth values in $A$ are such to make each rule whose head is $x$ unsatisfied by $A$. In other words, the direct connections with a node represent the \emph{support} for the node being in (or not in) the answer set under consideration.  
We refer the readers to the paper by Pontelli et al.\cite{PontelliSE09} for an in-depth discussion of properties of off-line justifications and the proof of existence of such justifications for every atom in the program.
    
Given a program $P$ , an answer set $A$ of $P$, and an atom $a$ occurring in $P$, explanation graphs for $a$ can be generated by (\emph{i}) determining the set of assumptions $U$ with respect to $A$; (\emph{ii}) generating explanation graphs for $a$ using $P$, $A$, and $U$ following its definition.   

\begin{figure} [ht]
   \begin{center}
   \begin{tabular}{c} 
   \includegraphics[height=5cm]{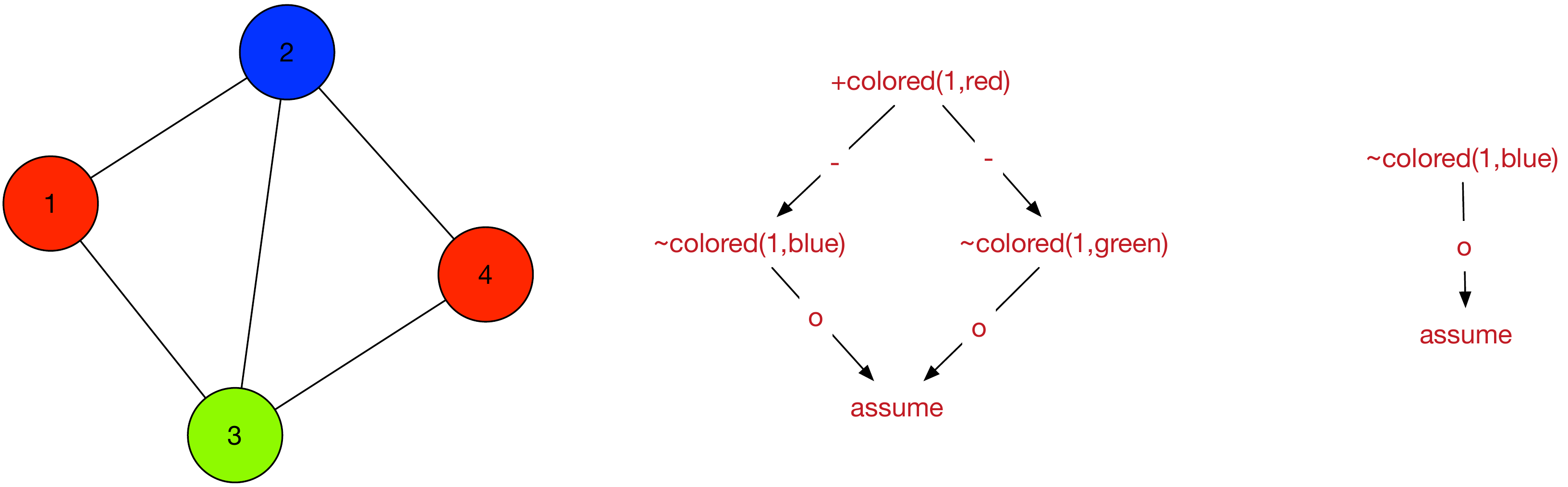}
   \end{tabular}
   \end{center}
   \caption[example] 
   { \label{fig:fig1} 
A solution to the 3-coloring problem for a graph on the (left), an explanation for positive atom $colored(1, red)$ (middle), and an explanation for negative atom $colored(1, blue)$ (middle)}
   \end{figure}

Fig. ~\ref{fig:fig1} illustrates the above definitions for the graph coloring program, given the graph  $G = (\{1,2,3,4\},\{(1,2), \newline (1,3),  (2,3),  (2,4), (3,4)\})$ and a solution on the left, represented by the answer set containing $\{colored(1,red), \newline colored(2,blue),  colored(3,green), colored(4,red)\}$.

\noindent
For our computation in the next section, we define a \emph{derivation path} of an atom $a$ as follows.

\begin{definition} 
[Derivation Path]
\label{def:dpath}
Given a program $P$, an answer set $A$, a derivation path of an atom $a$ is a directed graph $D_a = (N_a,E_a)$ where $N_a \subseteq {\{x \mid  x \in A\} \cup \{ \sim{x} \mid x \not\in A\}} \cup \{\top , \bot\}$ and $E_a \subseteq N_a \times N_a \times  \{+,-\}$, where $(x,y,z) \in E_a$ 
represents a link from $x$ to $y$ with the label $z$,  and satisfies the following conditions:  
\begin{itemize}
    \item if $\{(\sim\!x, \bot, +) \in E_a\}$ then there exists no rule in $P$ whose head is $x$; 
    \item there exists no $x,y$ such that $(\top, x, y) \in E_a$ and $(\bot, x, y) \in E_a$;
    \item for every $x \in N_a \cap A$ and $x$ is not a fact in $P$, 
    	\begin{itemize} 
    	\item there exists no $y \in N_a \cap A$ such that  $(x,y,-)$; 
    	\item there exists no $\sim\!y \in N_a \cap \{\sim\!u \mid u \not\in A\}$ such that  $(x,\sim\!y,+)$ belongs to $E_a$; 
    	\item if $X^+ = \{a' \mid (x, a',+) \in E_a\}$ and $X^- = \{a' \mid (x, \sim\!a',-) \in E_a\}$ then 
    	$X^+ \subseteq A$ and $X^- \cap A = \emptyset$ and there is a rule $r$ whose head is $x$ in $P$ such that  
    	$r^+ = X^+$ and $r^- = X^-$;
     	\item $D_a$ contains no cycle containing $x$.  
    	\end{itemize} 
    \item for every $\sim\!x \in N_a \cap \{\sim\!u \mid u \not\in A\}$, 
    	\begin{itemize} 
    	\item there exists no $y \in N_a \cap A$ such that  $(\sim\!x,y,+)$ belongs to $E_a$; 
    	\item there exists no $\sim\!y \in N_a \cap \{\sim\!u \mid u \not\in A\}$ such that  $(\sim\!x, \sim\!y,-)$  belongs to $E_a$; 
    	\item if $X^+ = \{a' \mid (\sim\!x,   a',-) \in E_a\}$ and $X^- = \{a' \mid (\sim\!x,  \sim\!a',+) \in E_a\}$ then 
    	$X^+ \subseteq A$ and $X^- \cap A = \emptyset$ and for every rule $r$ whose head is $x$ in $P$ we have 
    	 that $r^+ \cap X^- \ne \emptyset$ or 	$r^- \cap X^+ \ne \emptyset$;  
    	\item any cycle containing $\sim\!x$ in $D_a$ contains only node in $N_a \cap \{\sim\!u \mid u \not\in A\}$.
    	\end{itemize} 
\end{itemize}
\end{definition}

\noindent
 Note that derivation path is  different from the explanation graph. The purpose of derivation graph is to compute $U$ in Section~\ref{subsec:minimal_assumption}, in order to develop the explanation graphs in Section~\ref{subsec:asp_based_explanation_graph}.  

\section{\easp{}: A SYSTEM FOR GENERATING EXPLANATIONS FOR ASP-PROGRAMS}

In this section, we describe the algorithm implemented in the system \easp{} that generates explanations for ASP-programs (Section~\ref{subsec:preprocessing}, \ref{subsec:minimal_assumption}, and \ref{subsec:asp_based_explanation_graph}). Given a program $P$, the preprocessing steps in Section~\ref{subsec:preprocessing} are used to obtain  simplified ground rules from the answer set solver {\tt clingo} and  compute support sets for each rule. 
Given an answer set $A$ of program $P$, the algorithms in Section~\ref{subsec:minimal_assumption} compute  minimal assumption sets with respect to $A$.  
Algorithms in Section~\ref{subsec:asp_based_explanation_graph} use a minimal assumption set obtained from Section~\ref{subsec:minimal_assumption}  to provide the explanation graphs for an atom in $P$. For simplicity of the presentation, the discussion in this section assumes an arbitrary but fixed program $P$, unless otherwise specified.

\subsection{Preprocessing}
\label{subsec:preprocessing}

In this development, {\small \tt clingo} is utilized as an off-the-shelf tool. Given a program $P$, simplified ground rules of $P$ are computed via the combination of the {\small \tt --text} and {\small \tt --keep-facts}  options and an intermediate language $\mathit{aspif}$  \cite{kaminski2017tutorial}. First, by directly utilizing the  {\small \tt --text} and {\small \tt --keep-facts}  options, plain text format are obtained. Then, a set of facts $F$ is extracted from such plain text. Next, each fact $f \in F$ is modified to become an external statement ($\#external ~ f$) of the program. Facts in $P$ are replaced by the external statements, creating  a modified program $P'$. By doing this, we  prevent {\small \tt clingo} from simplifying rules when computing the  $\mathit{aspif}$ representation of $P'$ via its grounder, {\small \tt gringo} \cite{gebser2007gringo}. Let us illustrate this process using Example \ref{example:p1}. The $\mathit{aspif}$  statements are given in Listing \ref{lst:ex1}.

\begin{example} \label{example:p1}
Let us consider the program $P_1$ containing the rules:
\[\begin{array}{clclcclclcclcl} 
(r_1) & \texttt{a} & \uffa & \texttt{k}, \:\:not\:\texttt{b}. & \hspace{1cm} & (r_4) &\texttt{b} & \uffa & \:\:not\:\texttt{a}. & \hspace{1cm} &   (r_6) & \texttt{f} & \uffa & \texttt{e}, \:\:not\:\texttt{k}, \:\:not\:\texttt{c}.\\
(r_2) & \texttt{k} & \uffa & \texttt{e}, \:\:not\:\texttt{b}. & & (r_5) & \texttt{c} & \uffa & \texttt{k}. & \hspace{1cm} & (r_7) & \texttt{e}. \\
(r_3) & \texttt{c} & \uffa & \texttt{a}, \texttt{b}.
  \end{array}
\]

The program $P_1'$ is as follows:
\[\begin{array}{clclcclclcclcl} 
(r_1) & \texttt{a} & \uffa & \texttt{k}, \:\:not\:\texttt{b}. & \hspace{1cm} & (r_4) &\texttt{b} & \uffa & \:\:not\:\texttt{a}. & \hspace{1cm} &   (r_6) & \texttt{f} & \uffa & \texttt{e}, \:\:not\:\texttt{k}, \:\:not\:\texttt{c}.\\
(r_2) & \texttt{k} & \uffa & \texttt{e}, \:\:not\:\texttt{b}. & & (r_5) & \texttt{c} & \uffa & \texttt{k}. & \hspace{1cm} & (r_7) & \#external & \texttt{e}. \\
(r_3) & \texttt{c} & \uffa & \texttt{a}, \texttt{b}.
  \end{array}
\]
\end{example}

 A rule of the form \eqref{rule} in Sec.~\ref{sec:asp}, in $\mathit{aspif}$  format, is of the form: 
 \begin{center}
 $1 ~ H ~B$    
 \end{center}
where $H$ describes the head of the rule, starting with $0$, and has the form: 
 \begin{center}
    $0 ~ n ~ i_c$    
 \end{center}
  where $n = \{0,1\}$ is the number of head atoms and $i_c$ is an integer identifying the atom $c$. If the head is empty, $n = 0$ and $i_c = 0$.
  
 The  body of the rule is described by  $B$ and has the  form:
 \begin{center}
    $0 ~ k ~i_{a_1} ~ ... ~ i_{a_m} ~-i_{b_1} ~ ...~ ~-i_{b_n}$   
 \end{center}  
   where $i_{a_i}$'s and $i_{b_i}$'s are the integer identifiers of atoms $a_i$'s and $b_i$'s, respectively, and $k = n +m $. For instance, in Listing \ref{lst:ex1}, line 3 represents  rule $r_4$. 
   
    The lines starting with $4$ represent the mapping of atoms to their unique integer identifiers  and have the form:
 \begin{center}
    $4~m~a~1~i$    
 \end{center}
  where $m$ is the length in bytes of atom $a$ and $i$ is $a$'s identifier. For instance, line 9 in Listing \ref{lst:ex1} shows that $2$ is an integer identifying atom $b$.  
   
 Because we had changed facts to external statements in the modified program, facts are represented in lines starting with $5$ and have the form: 
 \begin{center}
    $5~i~2$ 
 \end{center}
 where $i$ is an integer identifier of a fact. For instance, Line 2 in Listing \ref{lst:ex1} (and Line 12) shows that atom $e$ is associated with the integer identifying $1$  is a fact.
 
 \begin{lstlisting}[language=clingo,caption=Representing $P_1'$ in $\mathit{aspif}$ format, label=lst:ex1, mathescape=true,xleftmargin=.3\textwidth, breaklines=true][htbp]
asp 1 0 0
5 1 2
1 0 1 2 0 1 -3
1 0 1 4 0 2 -2 1
1 0 1 3 0 2 -2 4
1 0 1 5 0 2 2 3
1 0 1 5 0 1 4
1 0 1 6 0 3 -5 -4 1
4 1 b 1 2
4 1 k 1 4
4 1 a 1 3
4 1 e 1 1
4 1 c 1 5
4 1 f 1 6
0
\end{lstlisting}

We implement \easp{} using Python. Thus, from now on, whenever we mention a dictionary, we refer to a Python data structure dictionary. For a dictionary $D$, we use $D.keys()$ and $D[k]$ to the set of keys in $D$ and the value associated with $k$ in $D$, respectively.

We use the $\mathit{aspif}$ representation of the modified program $P'$ to create a dictionary 
$D_P = \{h \mapsto B \mid h \in H,  B = \{body(r) \mid r \in P,  head(r) = h\} \}$. For each rule $r$, $body(r)$ consists of $r^+$ and $r^-$.  $D_P$ is then used as an input in Algorithm \ref{alg:preprocessing} to compute the dictionary $E_P = \{k \mapsto V \mid k \in \{a \mid a \in A\} \cup \{\sim\!a \mid a \notin A\},  V = \{support(k,r) \mid r \in P\}\}$.
Algorithm \ref{alg:preprocessing} first checks the truth value of atom $a \in H$ (recall that $H$ is the Herbrand base of the given program). If $a$ is true w.r.t $A$ (Line \ref{check_truth_value}), the function \emph{true\_atom\_processing} is used to compute the supported set for $a$, following the first property in Sect.~\ref{sec:properties_asp}. Otherwise, 
the function \emph{false\_atom\_processing} is used to compute the supported set for $a$, following the second property in Sect.~\ref{sec:properties_asp}.

\begin{algorithm}[!ht]
    \DontPrintSemicolon
    
    \KwInput{$D$ - dictionary of rules (this is $D_P$), $F$ - facts of the program, $A$ - an answer set}   
    
    $E \gets \{\emptyset \mapsto \emptyset\}$    \tcp*{Initialize an empty dictionary $E$: $E.key()=\emptyset$}
    
    \For{$a \in H$}
    {
        $bodies \gets D[a]$   
        
        \If{$a \in A$} 
        {\label{check_truth_value}
            
            $E[a] = []$     \tcp*{Initialize an empty list for the value stored with $a$ in $E$}
        
            \For{$body \in bodies$}
            {
                $E \gets true\_atoms\_processing(a,r^+,r^-,E,A,F)$ \label{true_atoms}
            }    
        } 
        \Else 
        {
            
            $E[\sim\!a] = []$       \tcp*{Initialize an empty list for the value stored with $\sim\!a$ in $E$}
        
            \For{$body \in bodies$}
            {
                $E \gets false\_atoms\_processing(a,r^+,r^-,E,A)$ \label{false_atoms}
            }  
        }
    }
    
    \KwRet $E$
    
    \;
    
    \textbf{function}  {$true\_atoms\_processing(a,r^+,r^-,E,A,F)$}
    
    \If{$r^+ \subseteq A \land r^- \cap A = \emptyset$ } 
    {\label{true_atom_check}
        $S \gets \{p \mid p \in r^+\} \cup \{\sim\!n \mid n \in r^-\}$  \label{true:support}
        
        \If{$a \in F \land r^+ = \emptyset \land r^- = \emptyset$}
        {
            $S \gets \{``\mathtt{T}"\}$  \label{true:fact}
        }
        
        Append $S$ to $E[a]$  \label{true:add_support}
    }
    
    \KwRet $E$
    
    \; 
    
    \textbf{function} {$false\_atoms\_processing(a,r^+,r^-,E,A)$}

\DontPrintSemicolon
    $L \gets [\{\{ \sim\!p \} \mid p \in r^+ \land p \notin A\} \cup \{\{n\} \mid n \in r^- \land n \in A\}]$  \label{false:supports}
    
    \If{$\sim\!a \notin E.keys()$}
    {
        $E[\sim\!a] \gets L$  \label{false:assign_support}
    }
    \Else
    { 
        $T = []$  \label{false:suports_start}   \tcp*{Initialize an empty list $T$}
        
        \For{$s \in L$}
        {
            \For{$e \in E[\sim\!a]$ }
            {
                $t \gets s \cup e$
                
                Append $t$ to a list $T$ \label{false:suports_end}
            }
        }
        
        $E[\sim\!a] \gets T$  \label{false:update_support}
    }
\KwRet $E$
\caption{$preprocessing(D,F,A)$}    
\label{alg:preprocessing}
\end{algorithm}

\begin{example} \label{example:e}
Let us reconsider the program $P_1$ of Example \ref{example:p1} which has an answer set $A = \{b,f,e\}$.  The output of Algorithm \ref{alg:preprocessing}, dictionary $E_{P_1}$,  is as follows:
\begin{lstlisting}[numbers=none]
      E = {  f : [{~k, ~c, e}]
             b : [{~a}]
            ~c : [{~k, ~a}]
            ~k : [{b}]
            ~a : [{~k}, {b}]
             e : [{T}]
          }
\end{lstlisting}
As can be seen from Example \ref{example:e}, $E_{P_1}$  contains supported sets for the keys in $E_{P_1}$.  For instance, atom $e$ in rule $r_2$ is not in any supported set for atom $b$, so $e$ does not appear in $E_{P_1}[b]$'s value.
\end{example}

We will also use the $\mathit{aspif}$ representation to compute $NANT(P)$ (Section~\ref{sec:explanation_graph}). Note that each negative atom is represented by a negative integer whose absolute value is in the symbol table. For instance, Line 3 in Listing \ref{lst:ex1} shows that the atom associated with integer $3$ is a negative atom, and Line 11 in Listing \ref{lst:ex1} shows that $3$ is integer identifying atom $a$; thus, for $P_1$, we have $a \in NANT(P_1)$, and $NANT(P_1) = \{a,b,c,k\}$. 

\subsection{Minimal Assumption Set}
\label{subsec:minimal_assumption}
 
 In this section, we compute all minimal assumption sets $U$ w.r.t $A$ that satisfy conditions of an assumption set in Sect.~\ref{sec:explanation_graph}. For every $u \in U$, $u$ is false in $A$ and does not  belong to the well-founded model and it must be assumed to be false to guarantee that cycles in the explanation graphs are acceptable \cite{PontelliSE09}. 
 
 \begin{algorithm}[!ht]
\DontPrintSemicolon
     \KwInput{C(P) - A cautious consequence of a program $P$, NANT(P) - A set of negative atoms in $P$, E - A dictionary computed in Algorithm \ref{alg:preprocessing}. }
    $TU = \emptyset$    \tcp*{Initialize an empty set $TU$}
    
    $TA_P(A) = \{a \mid a \in NANT(P) \land a \notin A \land a \notin C(P)\}$ \label{tentative_assumpton}
    
    $(T,DA) = derivation\_path(TA_P(A),E)$  \label{derivation}
  
    $D = dependency\_assumption(DA)$ \label{dependency}
    
    \For{$M \in D$}
    {
        $U \gets M \cup T$   \label{minimal_assumption_set}
        
        $TU \gets TU \cup \{U\}$
    }
    
    \KwRet $TU$
    
    \;
    
    \textbf{function} {$derivation\_path(TA,E)$}
    
        $DA \gets \{\emptyset \mapsto \emptyset\}$ and $T' \gets \emptyset$
       \tcp*{Initialize an empty dictionary $DA$ and an empty set $T'$}

    \For{$a \in TA$}
    { \label{check_a_start}
        $O \gets TA \setminus \{a\}$   \label{other_tentative_atoms}
        
        $L \gets E[\sim\!a]$   \label{get_supported_sets}
        
        \For{$S \in L$}
        {\label{loopS}
            $E_a \gets \{\emptyset \mapsto \emptyset\} $\tcp*{Initialize an empty dictionary $E_a$}
            
            $E_a[\sim\!a] \gets [S]$
            
            $E_a \gets get\_connection(S,E,E_a)$    \label{connection_to_s}
            
            $M \gets \{N_i \mid N_i = {k \mapsto V \mid V \in E_a[k]} \land \forall k \in E_a.keys(), k \in N_i.keys() \land ( k \mapsto V_1 \in N_i \land k \mapsto V_2 \in N_i) \Rightarrow V_1 = V_2 \}$   \label{loop_S_get_M}
            
            \For{$N_i \in M$}
            {\label{loopM}
                $V \gets []$, $R \gets \emptyset$, $C = \{\emptyset \mapsto \emptyset\}$ and $D = \emptyset$

                $(safe,D') \gets check\_derivation\_path(\sim\!a,N_i,O,V,R,C,D)$   \label{check_derivation}
                
                \If{$safe = True$}  
                {   \label{safe_derivation_start}
                    $T' \gets T' \cup \{a\}$
                    
                    
                    $DA[a] \gets D'$   \label{safe_derivation_end}
                    
                    \Go \ref{check_a_start}   \label{go}
                }
            }
        }
    }
  
    $T \gets TA \setminus T'$   \label{get_T}
    
    \KwRet $(T,DA)$
    
    \;
    
    \textbf{function} {$dependency\_assumption(DA)$} \label{func:da}

    $DC \gets \{J \mid J = \{u_1, u_2, ..., u_n \mid u_i \in DA.keys() \land  u_{i+1} \in DA[u_{i<n}]\} \land  u_1 \in D[u_n]\}$  \label{dependency_cycle}
  
    $B \gets \{\{j_1, ...,j_n\} \mid (j_1, ...,j_n) \in J_1 \times .. \times J_n \land n = \bigm|\!DC\!\bigm| \land J_i \in DC\}$  \label{set_atoms_breal_cycles}
    
    $min(B) \gets \{M \mid \forall C \in B, C \ne M \implies \bigm|\!M\!\bigm| \leq \bigm|\!C\!\bigm|\}$  \label{minimal_atoms_to_break}
    
    \KwRet $min(B)$
    
    \;
    
    \textbf{function} {$get\_connection(S,E,E_a)$}
  
    \For{$e \in S$}  
    {\label{get_connection_e_start}
        \If{$e \notin E_a.keys() \land e \in E.keys()$}
        {
            
            $E_a[e] \gets E[e]$   \label{get_connection_e_end}
            
            \For{$S_e \in E[e]$}
            {
                $E_a \gets get\_connection(S_e,E,E_a)$   \label{get_connection_recursive}
            }
        }
    }
  
    \KwRet $E_a$

\caption{$assumption\_func(C(P),NANT(P),E)$}
\label{alg:assumption_func}
\end{algorithm}
 
 Algorithm \ref{alg:assumption_func} shows the pseudo-code for computing $TU$ which consists of all minimal assumption sets $U$. The cautious consequence  $C(P)$ is computed using {\small \tt clingo}. $NANT(P)$ and $E$ are obtained in Sec.~\ref{subsec:preprocessing}.  First, $TA_P(A)$, called \emph{the tentative assumption set}, is computed in Line \ref{tentative_assumpton}. It is the set of negative atoms that are false in $A$ and do not belong to $C(P)$.

 $T$ and $DA$ are computed by the function \emph{derivation\_path}  (Line \ref{derivation})
where $T = TA_P(A) \setminus T'$. 
  $T$ is the set of atoms from $TA_P(A)$ and do not belong to $T'$ which is computed by the for-loop (Line~\ref{check_a_start}--\ref{go}). 
     For each $x \in T'$, there exists one derivation path of $x$ that either depends on other atoms in $TA_P(A)$ or satisfies the cycle conditions in the definition of explanation graph. 
  Intuitively, the atoms in $T$ must be assumed to be false and $DA$ contains the dependencies among atoms in $TA_P(A)$.

$D'$ is computed in Algorithm \ref{alg:check_derivation}. It checks whether a derivation path of an atom $a$ is acceptable and updates the set of dependency atoms in \emph{the tentative assumption set} for $a$. 
A derivation path is unacceptable if it has a cycle containing true atoms. This is verified by Algorithm \ref{alg:cycle}.  

The computation of sets of minimal atoms that break all cycles in \emph{the tentative assumption set} is then done by the function \emph{$dependency\_assumption(DA)$} (Lines~\ref{func:da}--\ref{minimal_atoms_to_break}).

\begin{algorithm}[!ht]
    \DontPrintSemicolon
    $I \gets \emptyset$    \tcp*{Initialize an empty set $I$}
    
    \If{$k \in N.keys()$}
    {
        $I \gets N[k]$
    }
    
    $V \gets V \cup \{k\}$  
    
    Append $k$ to list $R$
    
    \For{$i \in I$}
    { \label{check_derivation:forLoopI}
        $C[k] \gets i$
        
        $j \gets i.remove(i.index(0))$   \label{dependecy_start}
        
        \If{$i.index(0) = ``\sim" \land j \in O$}
        {
            $Da \gets Da \cup \{j\}$    \label{dependecy_end}
        }
        \Else{
            \If{$i \notin V$} 
            {  \label{check_derivation:cycle_start}
                \If{$check\_derivation\_path(i,N,O,V,R,C,Da)[0] = False$}
                { \label{recursive_func}
                    \KwRet (False,Da)
                }
            }
            \Else
            {
                \If{$i \in R$}
                {  \label{check_derivation:cycle_end}
                    \If{$i.index(0) = ``\sim"$} 
                    { \label{check_derivation:check_cycle_start}
                        \If{$cycle\_identification(C,i,i) = False$}
                        {
                           \KwRet (False,Da)   \label{check_derivation:check_cycle_end}
                        }
                    }
                    \Else
                    {
                        \KwRet (False,Da)
                    }
                }
            }
        }    
    }
    
    $l = R.pop()$
    
    \If{$l \in C.keys()$}
    {
        $C.pop(l)$
    }

    \KwRet (True,Da)
\caption{$check\_derivation\_path(k,N,O,V,R,C,Da)$}    
\label{alg:check_derivation}
\end{algorithm}

\begin{algorithm}[!ht]
    \DontPrintSemicolon
    $v \gets C[s]$
    
    \If{$v \ne e \wedge s.index(0) = ``\sim" \land v.index(0) = ``\sim"$ }{ 
    \KwRet $cycle\_identification(C,v,e)$   \label{cycle_recurive_edge}}
  
    \If{$v = e \wedge s.index(0) = ``\sim" \land v.index(0) = ``\sim"$ }{  
      \KwRet True}
    
      \KwRet False

\caption{$cycle\_identification(C,s,e)$}    
\label{alg:cycle}
\end{algorithm}
 
 \begin{example} \label{example:u}
 Let use reconsider the Examples \ref{example:p1} and \ref{example:e}. 
    \begin{itemize}
     \item For the program $P_1$, we have:
    \[\begin{array}{lcl}
    	{TA}_P(A) & = & \{c,a,k\} \\
      \end{array}
    \]
    
     \item As can be seen from $E$ in Example \ref{example:e}, $c$, $a$ and $k$ derive from other atoms in $\{k,a\}$, $\{k\}$ and $\{a\}$, respectively. Also, there is a cycle associated to $k$ and $a$. Thus, we have
     \[\begin{array}{lcl}
    	T & = & \emptyset \\
    	D & =  & \{\{a\}\},\{k\}\}
      \end{array}
    \]
    \item Two minimal assumption sets are as follows:
     \[\begin{array}{lcl}
    	U_1 & =  &  \{a\}\ \\
    	U_2 & =  &  \{k\}
      \end{array}
    \]
    \end{itemize}
\end{example}
 
\subsection{ASP-based explanation graph}
\label{subsec:asp_based_explanation_graph}

\begin{algorithm}[!ht]
\DontPrintSemicolon
    \For{$u \in U$}
    { \label{update_supported_sets_start}
     
        $E[\sim\!u] \gets [{``\mathtt{assume}"}]$     \label{update_supported_sets_end}
    }

    $a' \gets a$ if $a \in A$ and $a' \gets \sim\!a$ if $a \notin A$   \label{exp:get_truth_value}
    
    $L \gets E[a']$   \label{exp:get_supported_set}
        
    \For{$S \in L$}
    { \label{same_start}
        $E_a \gets \{\emptyset \mapsto \emptyset\}$ \tcp*{Initialize an empty dictionary $E_a$}
        
        $E_a[a'] \gets [S]$
        
        $E_a \gets get\_connection(S,E,E_a)$    \label{exp:connection_to_s}
        
        $M \gets \{N_i \mid N_i = {k \mapsto V \mid V \in E_a[k]} \land \forall k \in E_a.keys(), k \in N_i.keys() \land ( k \mapsto V_1 \in N_i \land k \mapsto V_2 \in N_i) \implies V_1 = V_2 \}$
        
        \label{same_end}
        
        \For{$N_i \in M$}
        {
            $V \gets []$, $R \gets \emptyset$, $C \gets \{\emptyset \mapsto \emptyset\}$, and an empty graph $G \gets (\emptyset, \emptyset)$
            
            Add node $a'$ to $G$
            
            $graph \gets get\_graph(a',G,N_i,V,R,C)$   \label{exp:graph}
            
            \If{$graph \ne False$}   
            { \label{exp:graph_start}
                Draw $graph$  \label{exp:graph_end}
            }
        }
    }

\;

    \textbf{function}  { $get\_graph(k,G,N,V,R,C)$}    \label{func-sgraph} 

    $I \gets \emptyset$ \tcp*{Initialize an empty set $I$}
    
    \If{$k \in N.keys()$}
    {
        $I \gets N[k]$
    }
    
    $V \gets V \cup \{k\}$  
    
    Append $k$ to list $R$
    
    \For{$i \in I$}
    {
        Add node $i$ to $G$
        
        $e \gets (k,i,sign)$ \label{sign_edge}
    
        Add edge $e$ to $G$    
        
        \If{$i \notin V$} 
        {
            \If{$get\_graph(i,G,N,V,R,C) = False$}
            {
                \KwRet False
            }
        }
        \Else
        {
            \If{$i \in R$}
            {
                \If{$i.index(0) = ``\sim"$} 
                { \label{check_cycle_start}
                    \If{$cycle\_identification(C,i,i) = False$}
                    {
                       \KwRet False   \label{check_cycle_end}
                    }
                }
                \Else
                {
                    \KwRet False
                }
            }
        }
    }
    
    $l = R.pop()$
    
    \If{$l \in C.keys()$}
    {
        $C.pop(l)$
    }

    \KwRet G \label{func-ggraph} 

\caption{$explanation\_graph(a,E,U)$}
\label{alg:explanation_graph}
\end{algorithm}

In this section, we utilize the dictionary $E_P$ and a minimal assumption set $U$ to provide the explanation graphs of an atom $a$ in the program $P$. Algorithm \ref{alg:explanation_graph} shows the pseudo-code for computing the explanation graphs for $a$. Elements in $u\in U$ is $\{``\mathtt{assume}"\}$ (Lines \ref{update_supported_sets_start}--\ref{update_supported_sets_end}). Supported sets of $a$ are then computed and assigned to $L$ (Line \ref{exp:get_supported_set}). Lines \ref{same_start}--\ref{same_end} create $M$, a collection of dictionaries of derivation paths for $a$. It is then used for computing explanation graphs for $a$ via the function  \emph{get\_graph} (Lines \ref{func-sgraph}--\ref{func-ggraph}).
If the derivation graph is an explanation graph, it will be displayed as a graph on the screen.

\begin{example}
For program $P_1$ in Example \ref{example:p1}, the explanation graphs of $f$ w.r.t $U_1$ and $U_2$ are shown on the left and right of Fig. \ref{fig:f}, respectively. 

As can be seen from Fig.~\ref{fig:f}, a justification for $f$ depends negatively on $c$ and $k$, and positively on $e$. Based on the minimal assumption set we choose, we receive two different explanation graphs for $f$. The left figure contains $b$ in its graph while the right figure does not. 

\begin{figure}[h!]
  \begin{center}
  \begin{tabular}{c}
  \includegraphics[width=.25\textwidth]{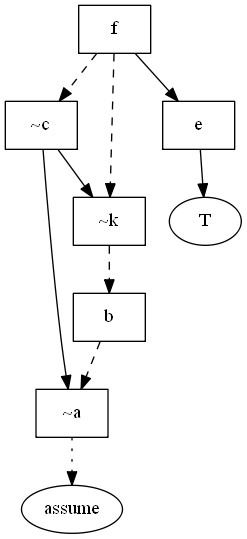}
  \quad \quad \quad \quad 
  \includegraphics[width=.25\textwidth]{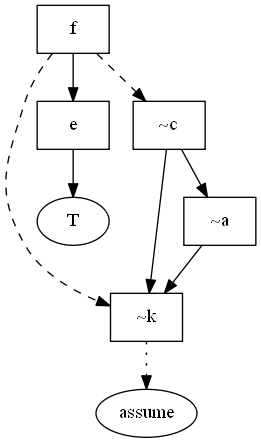}
  \end{tabular}
  \end{center}
  \caption{\scriptsize Explanation graph of $a$ w.r.t $U_1$ (left) and $U_2$ (right)}
  \label{fig:f}
\end{figure}
\end{example}

\section{ILLUSTRATION}

We have validated a prototype implementation of this system on a number of ASP programs, ranging from simple benchmarks to complex ASP applications (e.g., we used it to interpret the responses produced by a diagnosis system modeling the  Three Mile Island disaster \cite{hanna2020application}).
In this paper, we illustrate the use of \easp{} in two  examples from the literature.  

\begin{example} \label{example:bob}
 Consider the following problem, which represents a variant of the agent problem described by Garcia et al.\cite{garcia2013formalizing}. An agent, named Bob,  wants to make plan for the upcoming week. Bob has two choices: stay home or go to the opera. Bob will stay at home on Monday. He will also stay at home with his best friend because she has a baby, and the friend comes this Tuesday. The situation can be encoded as follows:
\[\begin{array}{lclclclclcl}
\texttt{day(monday)}. & \hspace{.5cm} & \texttt{day(tuesday)}. & \hspace{.5cm} & \texttt{day(wednesday)}. \\
\texttt{day(thursday)}. & \hspace{.5cm} & \texttt{day(friday)}. & \hspace{.5cm} & \texttt{day(saturday)}. \\
\texttt{day(sunday)}. & \hspace{.5cm} & \texttt{home(monday)}. & \hspace{.5cm} & \texttt{baby(tuesday)}. \\
\texttt{opera(D)} & \uffa & \texttt{day(D)}, \:\:not\:\texttt{ home(D)}. \\ 
\texttt{home(D)} & \uffa & \texttt{day(D)}, \:\:not\:\texttt{ opera(D)}.\\
\texttt{home(D)} & \uffa & \texttt{day(D)},\texttt{ baby(D)}. \\
\end{array}
\]

The above program has 32 answer sets, i.e., Bob has 32 possible plans. Let us consider  the answer set   $A = \{day(monday),day(tuesday), day(wednesday), day(thursday), day(friday), day(saturday), day(sunday), \\ home(monday),  home(tuesday), baby(tuesday), opera(wednesday), opera(thursday), opera(friday), \\ opera(saturday),  opera(sunday)\}$. 

Some explanation graphs w.r.t $A$ are displayed below:
\begin{figure}[h!]
  \begin{center}
   \begin{tabular}{c} 
      \includegraphics[width=.15\textwidth]{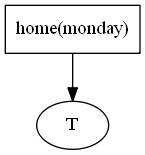} \quad 
      \includegraphics[width=.30\textwidth]{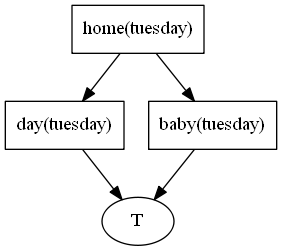} \quad 
      \includegraphics[width=.30\textwidth]{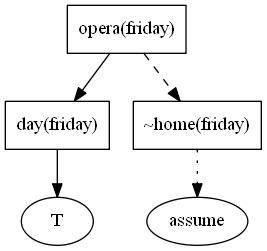}
  	\end{tabular}
	\end{center}
  \caption{\scriptsize Explanation of $home(monday)$ (left), $home(tuesday)$ (middle) and  $opera(friday)$ (right)}
  \label{fig:example1}
\end{figure}
\begin{itemize}
    \item Fig.~\ref{fig:example1} (left) shows that Bob will stay at home on Monday because of the provided information in the situation that $home(monday)$ is a fact.
    \item Fig.~\ref{fig:example1} (middle) shows that Bob will stay at home on Tuesday because his best friend and her baby will come to his house on Tuesday ($baby(tuesday)$ is a fact).
    \item Fig.~\ref{fig:example1} (right) shows that Bob will go to opera on Friday because it is assumed to be false that he will not stay at home on Friday.
\end{itemize}
\end{example}

\begin{example}  \label{example:treatment}
  Consider the problem of decision making in an ophthalmologist diagnostic system \cite{schulz_toni_2016}. We  need to provide a suggestion to Peter. Peter is short-sighted. He is afraid to touch his eyes. He is a student and likes sport. The information about Peter is encoded by the following facts.

\[\begin{array}{lclclcl}
\texttt{shortSighted}. & \hspace{.5cm} & \texttt{afraidToTouchEyes}. \\
\texttt{student}. & \hspace{.5cm} & \texttt{likesSports}.  \\
\end{array}
\]
 Other information about Peter is given by the following program:
\[\begin{array}{lclclcl}
\texttt{tightOnMoney} & \uffa & \texttt{student}, \:\:not\:\texttt{richParents}. \\ 
\texttt{caresPracticality} & \uffa & \texttt{likesSports}.\\
\texttt{correctiveLens} & \uffa & \texttt{shortSighted},  \:\:not\:\texttt{laserSurgery}. \\
\texttt{laserSurgery} & \uffa & \texttt{shortSighted},  \:\:not\:\texttt{tightOnMoney},  \:\:not\:\texttt{correctiveLens}. \\
\texttt{glasses} & \uffa & \texttt{correctiveLens},  \:\:not\:\texttt{caresPracticality}, \\
& &\:\:not\:\texttt{contactLens}. \\
\texttt{contactLens} & \uffa & \texttt{correctiveLens},  \:\:not\:\texttt{afraidToTouchEyes},
\\
& &\:\:not\:\texttt{longSighted}, \:\:not\:\texttt{glasses}
. \\
\texttt{intraocularLens} & \uffa & \texttt{correctiveLens},  \:\:not\:\texttt{glasses},  \:\:not\:\texttt{contactLens}. \\
\end{array}
\]

\noindent
This program, together with the facts,  has an answer set 

$A = \{tightOnMoney,shortSighted,caresPracticality,afraidToTouchEyes,
 student, likesSports, \\ correctiveLens,intraocularLens\}$. 

\noindent
Therefore, $intraocularLens$ is suggested to Peter. The explanation graph for this recommendation is depicted in Fig.~\ref{fig:example2}. 
\begin{figure}[htbp]
  \centering
  \includegraphics[width=.50\textwidth]{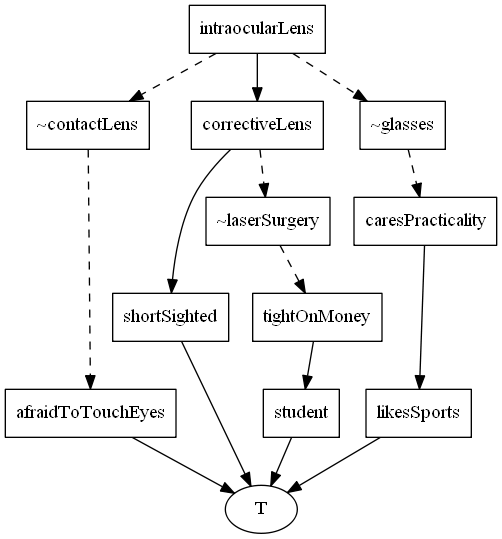}
  \caption{\scriptsize Explanation of  $intraocularLens$}
  \label{fig:example2}
\end{figure}

As can be seen from Fig.~\ref{fig:example2}, $intraocularLens$ is in the answer set because:
\begin{itemize}
    \item The truth value of $correctiveLens$ is True, because
    \begin{itemize}
        \item It has information that Peter is short-sighted ($shortSighted$ is a fact).
        \item Negatively it depends on the truth value of $laserSurgery$ which is False. This latter truth value derives from the fact that $laserSurgery$ depends negatively on  the truth value of $tightOnMoney$ which is True. This is  because it has information that Peter is student ($student$ is a fact).
    \end{itemize}
     \item The truth value of $contactLens$ is False, because
     \begin{itemize}
         \item Negatively it depends on the truth value of $afraidToTouchEyes$ which is a fact.
     \end{itemize}
    \item The truth value of $glasses$ is False, because
     \begin{itemize}
         \item Negatively it depends on the truth value of $caresPracticality$ which is True, because
         \begin{itemize}
             \item It has information that Peter likes sports ($likeSports$ is a fact).
         \end{itemize}
     \end{itemize}
\end{itemize}
\end{example}

\section{CONCLUSION} 

In this paper, we described an explanation generation system for ASP programs, \easp{}, and illustrate its use in some examples. Our  future goal is to use the proposed system in explainable planning with the full cycle of four steps (1)-(4) in Fig.~\ref{fig0}. 

\acknowledgments 
 
This research is partially supported by NSF grants 1757207, 1914635, and 1812619/26. Portions of this publication and research effort are made possible through the help and support of NIST via cooperative agreement 70NANB19H102. The views and conclusions contained in this document are those of the authors and should not be interpreted as representing the official policies, either expressed or implied, of the sponsoring organizations, agencies, or the U.S. government.  

\bibliography{report} 
\bibliographystyle{spiebib} 

\end{document}